\documentclass[12pt,a4paper]{article}

\pdfoutput=1

\usepackage[cmex10,intlimits]{amsmath}
\usepackage{wasysym}
\usepackage{amssymb}
\usepackage{amsthm}
\usepackage{amsfonts}
\usepackage{amsmath}
\usepackage{color}
\usepackage{a4wide}
\usepackage{rotating}
\usepackage{cite}

\makeatletter
\def\blfootnote{\xdef\@thefnmark{}\@footnotetext}
\makeatother

\title{Scale-Regularized Filter Learning:\\ Calculus of Variation meets Learning}

\author{Marco~Loog \medskip \\
\small
\begin{tabular}{r|l}
Pattern Recognition Laboratory & The Image Section \\
Delft University of Technology & University of Copenhagen \\
The Netherlands & Denmark \\
e-mail: m.loog@tudelft.nl & http: prlab.tudelft.nl
\end{tabular} \bigskip \\
Fran\c{c}ois~Lauze \medskip \\
\small
\begin{tabular}{r|l}
 & The Image Section \\
 & University of Copenhagen \\
 & Denmark \\
e-mail: lauze@di.ku.dk & http: www.diku.dk
\end{tabular}}

\date{December 15, 2014\footnote{Original submission to SSVM 2015 with a few minor corrections. Version with minor changes to appear in Proceedings of the BMVC 2017 as \emph{Supervised Scale-Regularized Linear Convolutionary Filters}.}}

\begin{document}


\maketitle

\begin{abstract}
We start out by demonstrating that an elementary learning task, corresponding to the training of a single linear neuron in a convolutional neural network, can be solved for feature spaces of very high dimensionality.  In a second step, acknowledging that such high-dimensional learning tasks typically benefit from some form of regularization and arguing that the problem of scale has not been taken care of in a very satisfactory manner, we come to a combined resolution of both of these shortcomings by proposing a form of scale regularization.  Moreover, using variational method, this regularization problem can also be solved rather efficiently and we demonstrate, on an artificial filter learning problem, the capabilities of our basic linear neuron.  From a more general standpoint, we see this work as prime example of how learning and variational methods could, or even should work to their mutual benefit. \medskip \\
{\bf Keywords}: regression, learning methods, scale, variatonal methods, convolutional neural networks.
\end{abstract}


%

\newpage

\section{Introduction}

Nowadays, many computer vision and image analysis tasks are tackled by means of pattern recognition and machine learning techniques.  This work makes some initial steps in the opposite direction.  It does not reject the learning approach to computer vision, but it shows how tools form computer vision---and especially variational methods, can aid in efficiently solving some of the basic estimation tasks machine learners and pattern recognizers come across.  The particular issue we consider is the problem of scale that, one way or the other, emerges in any learning tasks involving images or videos. As so often, however, it is overlooked and/or dealt with in a way that leaves much to be desired.  This work is really at the very interface of computer vision and learning techniques and we may draw heavily on terminology from both fields.  The main concept from machine learning and pattern recognition that we use are learning (or training) from examples (especially in relation to linear regression) and the ideas underlying artificial neural network, and convolutional neural in networks particular (see, for instance, \cite{bishop2006pattern,hastie01} and \cite{bishop1995neural,lecun1989backpropagation}).   Nonetheless, we think that researchers schooled in scale space and variational methods should be able to follow our main line of thought.

\subsection{Pixel-based Classification and Regression}

Supervised classification and regression techniques have been applied to a broad range of challenging image processing and analysis tasks.  Learning-based pixel classification has been around at least since the 1960s.  Early studies seem to have been conducted particularly within the field of remote sensing and abutting areas \cite{fu1969information}.  Though these approaches initially seemed to have focussed primarily on the use of the multiple spectral bands that the observations consisted of, later work also include spatial features based on derivative operator, texture measures, and the like (cf. \cite{nagy1972digital}).  An early overview of the general applicability of pixel classification can be found, for instance, in \cite{holdermann1978review}.

Training image filters on the basis of given input-output pairs by means of regression techniques seems to have been considered less often.  The problem, as opposed to pixel classification, may be obtaining proper input and output image pairs.  Also for this reason, possibly the most often studied application is the prediction of (supposedly) noiseless images from images corrupted with a known noise model, as in this case the input-output pairs are easily generated.  Perhaps the first paper to consider such option is \cite{hou2004image}---but the approach may only turn popular now it has been presented at a more fashionable venue \cite{burger2012image}.  A few years later, more advanced applications found their way into medical image analysis, in particular for filtering complex image structures out of chest radiographs \cite{loog2006filter,suzuki2006image}.

Where not so many years ago, pixel-based methods relied on features extracted by means of more or less complex, linear or nonlinear filter banks, the past decade has seen a trend of so-called representation learning \cite{bengio2013representation}.  The idea is to avoid any initial (explicit) bias in the learning that entails from prespecifying the particular image features that are going to be used.  Rather, one relies on raw input data (images in our case) and a complex learner that is capable of simulating the necessary filtering based on the raw input.  Particular architectures that are used as learners are so-called deep networks, which are simply a specific type of artificial neural networks.  Two examples of the use of such networks in image denoising can be found in \cite{jain2009natural} and the earlier mentioned work \cite{burger2012image}.  A first approach to supervised segmentation using these methods can be found in \cite{grangier2009deep}.


We see the current work in the light of these developments in representation learning, although our results are not ``deep''.  In fact, here we will deal with shallow networks with a single linear convolutional neuron \cite{bengio2013representation}; a basic element in the more complex deep structures referred to above. The problem we focus on is inferring an image-to-image mapping, which is not necessarily limited to image denoising.  The core of the issue we study is how to control the complexity of that single neuron. In our case, this is achieved by controlling the aperture scale at which the neuronal mapping operates.  In current applications of deep networks, notably convolutional neural networks \cite{lecun1989backpropagation}, the spatial extent from which the different layers draw their information is coarsely modeled by a rectangle with preset dimensions.  We propose to not prefix the spatial range explicitly---and to basically have every pixel intensity in every location have potential influence on any other pixel. We decide to integrate the influence of scale by a regularization term into the overall objective function that is used to determine the fit of the neuron to the training data.  In this way, we can trade off the influence of the training data and the scale of the aperture in a gradual and controlled way.

\subsection{Outline}

Section \ref{sect:prob} formulates the initial problem setting in mathematical terms.  The loss on the data term considered is the regular squared error and so we are basically dealing with standard least squares linear regression.  The section shows that our nonregularized prediction problem can be seen as a convolution in which the convolution kernel is to be determined.  As it turns out, the formulations allows us to solve regression problems in features spaces with very high dimensionality\footnote{As we are in the setting of representation learning, features are pixels values here.} and with even larger numbers of observations.  Section \ref{sect:reg}, covering the main part of our theory, argues that some form of regularization would typically be necessary, after which it introduces and explains our scale-regularized objective function.  It also shows how to reformulate the optimization problem so that its minimization can be performed by means of a variational method and finally sketches a basic scheme to come to an actual solution.  Section \ref{sect:exp} provides some limited and artificial, yet illustrative examples and Section \ref{sect:conc} discusses and concludes.

\section{Regression and Supervised Filter Learning}\label{sect:prob}

Let us initially considering a set of $N$ input-output pairs of images $\{(\alpha_i,\beta_i)\}_{i=1}^N$ defined on the full image domain $\mathbb{R}^d$: $\alpha_i, \beta_i:\mathbb{R}^d \to \mathbb{R}$.  We do not consider multi-band or multi-spectral images, but out theory is equally applicable to this setting.  Given these $N$ pairs, of what we will refer to as training images, we would like to infer a transformation $T$ that can be applied to any new and unseen input image $\alpha$, such that it optimally predicts its associated, and unobserved, output $\beta$.  The expected least squares loss $L$ between the true output and the prediction by $T$ is typically used to define optimality of the transformation $T$:
\begin{equation}\label{eq:gen}
L[T] = \int p(\alpha,\beta) \int | T[\alpha](x) - \beta(x) |^2 \, dx \, d\alpha \, d\beta \, ,
\end{equation}
where we tacitly assume the integrals exist.  In the absence of any precise knowledge of $p$, the prior over pairs of input and output images, the true expected loss must be approximated.  If there is training data available we may rely on the empirical risk, which is determined by substituting the empirical distribution of our observations for $p$, leading to the objective
\begin{equation}\label{eq:emp}
L[T] = \frac{1}{N} \sum_{i=1}^N \int | T[\alpha_i](x) - \beta_i(x) |^2 \, dx \, .
\end{equation}

In many a setting, the transformation $T$ would be taken translation invariant.  This is the situation we consider here as well.  In fact, since we focus on a single linear convolutional neuron, $T$ reduces to a simple linear convolution by means of a kernel $u:\mathbb{R}^d \to \mathbb{R}$. Equation (\ref{eq:emp}) therefore simplifies to
\begin{equation}\label{eq:lin}
L[u] = \frac{1}{N} \sum_{i=1}^N \int | (\alpha_i \ast u)(x) - \beta_i(x) |^2 \, dx \, .
\end{equation}
Denoting the Fourier transform by $\mathcal{F}$ or $\hat{ }$, the optimal solution $K^\star$ to the above equation can be obtained as
\begin{equation}\label{eq:opt}
u^\star = \mathcal{F}^{-1} \left[ \frac{\sum_{i=1}^N \hat{\beta}_i\bar{\hat{\alpha}}_i }{\sum_{i=1}^N \hat{\alpha}_i\bar{\hat{\alpha}}_i } \right] \, .
\end{equation}
This formulation, in fact, allows us to efficiently solve an image regression problem in, potentially, very high dimensional feature spaces.  To see that Equation (\ref{eq:lin}) basically formulates a regular linear regression problem, note that $(\alpha \ast u)(x)$ equals $\langle \alpha(x - \cdot), u \rangle$, where $\alpha(x - \cdot)$ are the explanatory variables or the feature ``vectors'' indexed by the variable $x$ and $u$ can be interpreted as an estimate for the true regression parameters.  Indeed, instead of using patches of limited size to capture the contextual information around every pixel location, this formulation basically takes the whole image (centralized around $x$) to be the patch to every location $x$.

\subsection{Regression Problem Size, an Example}

\begin{figure}[!ht]
\centering
\includegraphics[height=6cm]{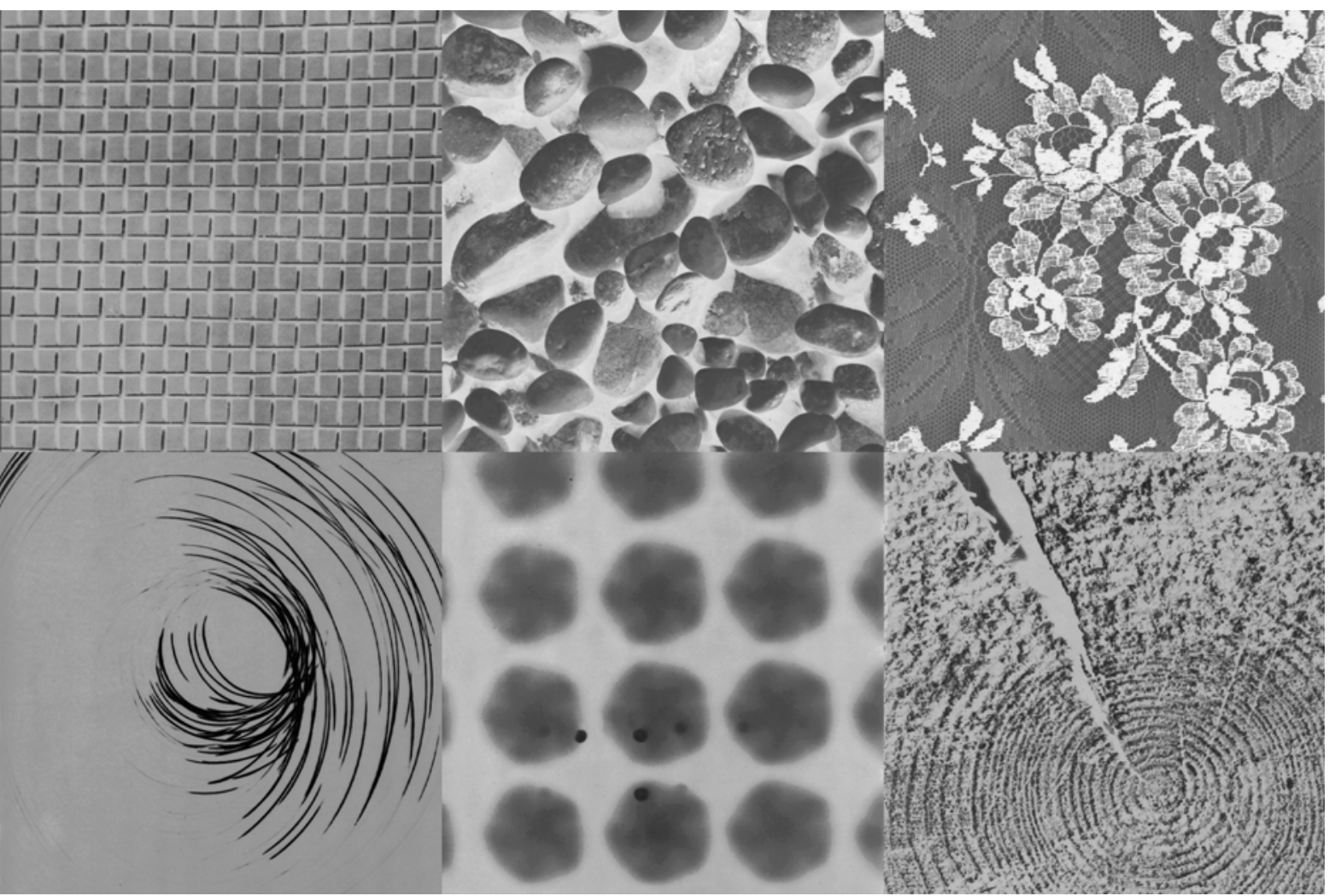}
\caption{Example input images from the Brodatz data set.}\label{fig:in}
\end{figure}

Consider the Brodatz images data set that we are going to experiment with later on \cite{brodatz1966textures}.  The set consists of 112 images of dimensions $640 \times 640$, which we take as the input images (some examples are shown in Figure \ref{fig:in}).
Let us assume that we have corresponding output images to all of the 112 Brodatz images, which are the original images corrupted by an unknown convolution filter and additive Gaussian noise.  Finding a filter $u$ that is optimal in the empirical least squares sense means that one would actually have to solve a linear regression problem in $640^2 > 400$ thousand dimensions, coming from the patch size we consider. The number of instances, we would base the learning on is $640^2 \times 112$, which is more than $45$ million, as there are $640^2$ locations per image and we have $112$ images.

Solving this problem in the standard way by means of linear regression would, among others, mean that we have to invert a covariance matrix sized $640^2 \times 640^2$, which is sheer impossible.  Because of the convolutional structure of the problem, however, explicit matrix inversion can be avoided and the computationally most demanding part in Equation (\ref{eq:opt}) is the Fourier transformation.  Relying on the fast Fourier transform, the necessary computations to find the more than 400 thousand weights of our neuron (encoded through $u$) can be done in one or two seconds, even on a modest laptop.

\section{Scale Regularization}\label{sect:reg}

Using basic image processing techniques, for every pixel location in an image, we can actually include all other image values as context in its feature vector and still solve the high-dimensional regression problem efficiently.  Nonetheless, there is a good reason why a convolutional neural network would restrict the extent of every filter to an area considerably smaller than the whole image.  Estimations in such high-dimensional spaces easily leads to overtraining or overfitting as there are too many free parameters to be estimated compared to the number of observations that may be available.  But it seems reasonable to assume that it is more likely that the useful predictive information for a particular location in an image comes from locations nearby rather then pixel values far away.

The current way to exploit this kind of prior knowledge is by explicitly extracting patches of limited size around every pixel location and base the regression on these features only.  Equivalently, the convolutional objective function in Equation (\ref{eq:lin}) can be adapted to do the same by simply restricting the support of the filter $u$, i.e., one can minimize for Equation (\ref{eq:lin}) under the constraint that $\mathrm{supp} \, u$ is an appropriate subset $\Omega$ of $\mathbb{R}^d$.  Typically, $\Omega$ is just taken to be a square patch.

Here, we suggest to take care of scale in, what we think is a more proper way.  Instead of restricting the influence of surrounding pixel values to a particular region explicitly, we propose to gradually suppress the influence of more and more distant pixel values by means of a scale-sensitive regularization term on the kernel $u$, which we add as a term to our original least squares objective function in Equation (\ref{eq:lin}).  In particular, we consider minimizing the following:
\begin{equation}\label{eq:reg}
L[u] = \frac{1}{N} \sum_{i=1}^N \int | (\alpha_i \ast u)(x) - \beta_i(x) |^2 \, dx + \lambda \int ||x||^2 u^2(x) \, dx \, ,
\end{equation}
where $\lambda \ge 0$ controls the scale.

The primary characteristic of the regularizing term is that larger values for $K$ should be discouraged the further away one gets from the center of the kernel.  Clearly, various other formulations would have been possible, but the current suggestion has some appealing properties.  Firstly, the polynomial $||a||^2$ is rotationally invariant.  Secondly, it is homogenous, so changing the unit in which we measure distance to the kernel center, can equivalently be accommodated by changing $\lambda$, i.e., the effect of substituting $||ca||$ for $||a||$, can also be achieved by substituting $c^2\lambda$ for $\lambda$.  Still, such properties would hold for any choice of power, not only for the square.  Choosing the square, however, leads to a relatively easy to solve variational problem, which allows us to retain some of the computationally attractive properties of the original formulation in (\ref{eq:lin}).

\subsection{Minimization } \label{sect:min}

The choice of the regularization term in (\ref{eq:reg}) makes the minimization easy via Fourier
Transform: using the derivation properties of the Fourier transform as well as Plancherel's theorem,
one gets for an $L^2(\mathbb{R}^d)$ function $u$:
\begin{eqnarray}
\|\nabla u\|^2 &=& \sum_{j=1}^d \|\partial_{x_j}u\|^2 = \sum_{j=1}^d \int_{\mathbb{R}^d} |\partial_{x_j}u(x)|^2dx\\
&=& 4\pi^2\sum_{j=1}^d \int_{\mathbb{R}^d}\xi_j^2 \hat{u}^2(\xi)\,d\xi = 4\pi^2\int_{\mathbb{R}^d}||\xi||^2\hat{u}^2(\xi)\,d\xi \, .
\end{eqnarray}
Using the properties of the convolution and the Fourier transform, we can rewrite the criterion in Equation (\ref{eq:reg}) as
\begin{equation}
  L[\hat{u}] = \frac{1}{N} \sum_{i=1}^{d}\|\hat{\alpha}_i \hat{u} - \hat{\beta_i}\|^2 + \frac{\lambda}{4\pi^2}\|\nabla \hat{u}\|^2 \, ,
\end{equation}
which is a Tikhonov regularization of the regression problem.  By letting $a = (\hat{\alpha}_1,\dots,\hat{\alpha}_n)^\top$ and $b = (\hat{\beta}_1,\dots,\hat{\beta}_n)^\top$, we can rewrite it as
\begin{equation}
  \label{eq:regfourier}
   L[\hat{u}] = \frac{1}{N} \|\hat{u} a - b\|^2 + \frac{\lambda}{4\pi^2}\|\nabla \hat{u}\|^2.
\end{equation}
Computing the first variation of Equation (\ref{eq:regfourier}) gives the optimality condition
\begin{equation}
  \label{eq:optim}
  a^\ast (a\hat{u} - b) - \frac{\lambda}{4\pi}\Delta \hat{u} = a^\ast a\hat{u} - a^\ast b - \frac{\lambda}{4\pi}\Delta \hat{u} = 0 \, ,
\end{equation}
where $a^\ast$ denotes the Hermitian adjoint of $a$, $a^\ast = {\bar{a}}^\top$. Note that because the $\hat{\alpha}_i$ and $\hat{\beta}_i$ are Fourier transforms of real functions they are Hermitian, i.e., they satisfy the equation
\begin{equation}
\label{eq:conjcomp}
\hat{f}(-\xi) = \bar{\hat{f}}(\xi) \, .
\end{equation}
Of course the solution will be Hermitian as well, as is easily checked.

More importantly from a computational perspective, note that $a^\ast a = \sum \alpha_i^\ast \alpha_i$ and $a^\ast b = \sum \alpha_i^\ast \beta_i$ are, next to the value of $\lambda$, the only inputs to the optimization one needs. The size of both $a^\ast a$ and $a^\ast b$ equals the original image size and does not depend on the number of training images. This makes explicit that no matter how many training images one uses, once we have  $a^\ast a$ and $a^\ast b$, the computational complexity of getting to a solution for Equation (\ref{eq:optim}) remains the same.

Now, the actual numerical minimization for the 2-dimensional images in our experiment is carried out using a standard 5-points stencil for the Laplacian, using periodic boundary conditions. The resulting system is solved by Jacobi relaxation and reads as follows:
\begin{equation}\label{eq:jacobi}
\hat{u}^{n+1}_{ij} = \frac{(a^\ast b)_{ij} +
  \frac{\lambda}{4\pi^2}\left(\hat{u}^{n}_{i+1j}+\hat{u}^{n}_{i-1j}+\hat{u}^{n}_{ij+1}+\hat{u}^{n}_{ij-1}\right)}{(a^\ast a)_{ij}
  + \frac{\lambda}{\pi^2}}
\end{equation}
(omitting boundary conditions).  Though faster solvers are possible, the use of the Jacobi solver automatically enforces at each iteration the discrete counterpart of the Hermitian relation in Equation (\ref{eq:conjcomp}) and therefore, at each iteration, $\hat{u}^n$ remains the Fourier transform of a real signal.  Finally note that in Equation \ref{eq:jacobi}, the regularizing effect of a positive $\lambda$ can, in part, be seen back, as it keep the denominator bounded away from zero.

\begin{figure}[!ht]
\centering
\includegraphics[height=4cm]{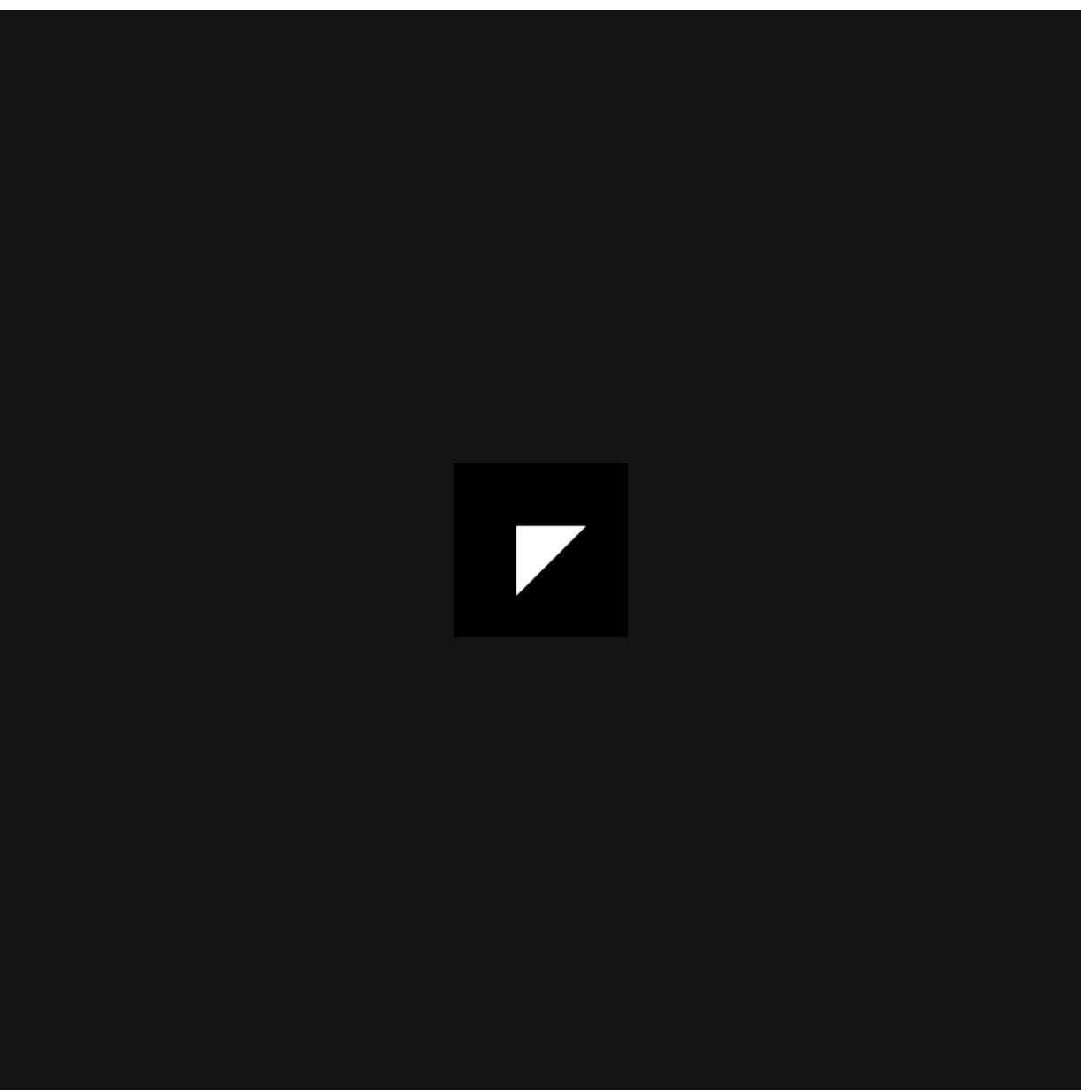}
\includegraphics[height=4cm]{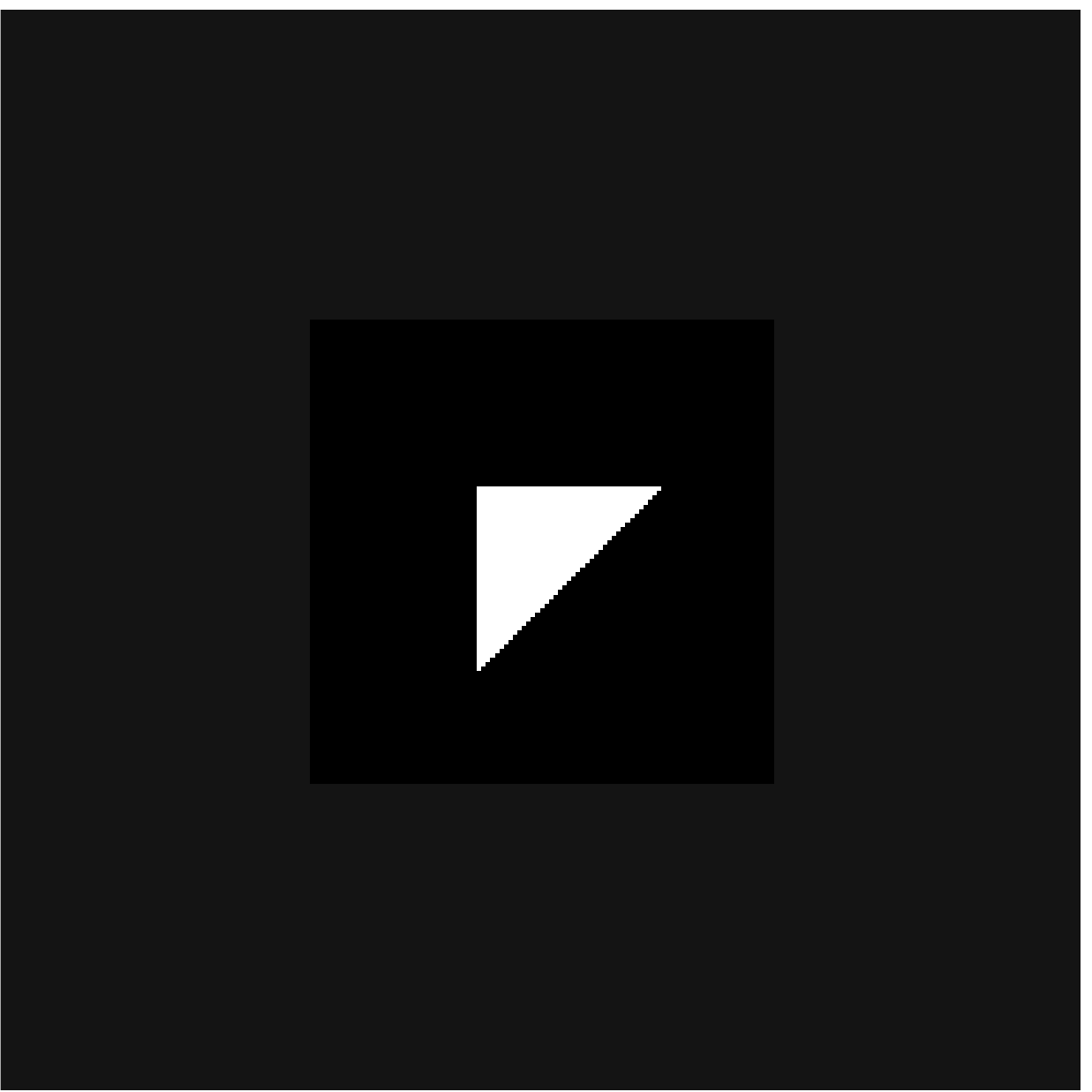}
\caption{On the left: convolution kernel used to create output images from the original Brodatz images.  On the right: a cropped version for future comparison.}\label{fig:Forg}
\end{figure}
\begin{figure}[!ht]
\centering
\includegraphics[height=6cm]{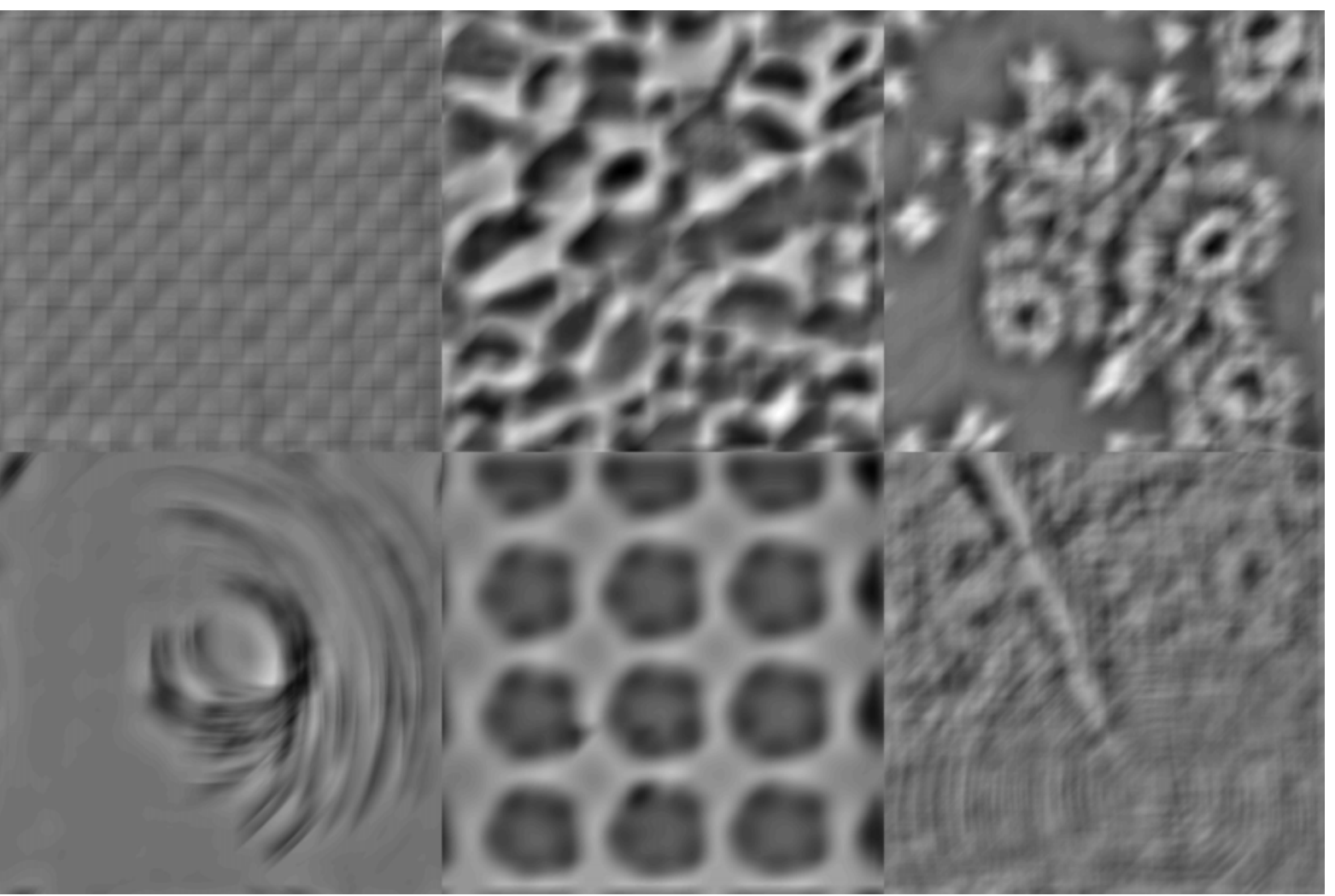}
\caption{Examples of initial output images obtained by convolving the input images from Figure \ref{fig:in} with the kernel depicted in Figure \ref{fig:Forg}.}\label{fig:nonoise}
\end{figure}
\begin{figure}[!ht]
\centering
\includegraphics[height=6cm]{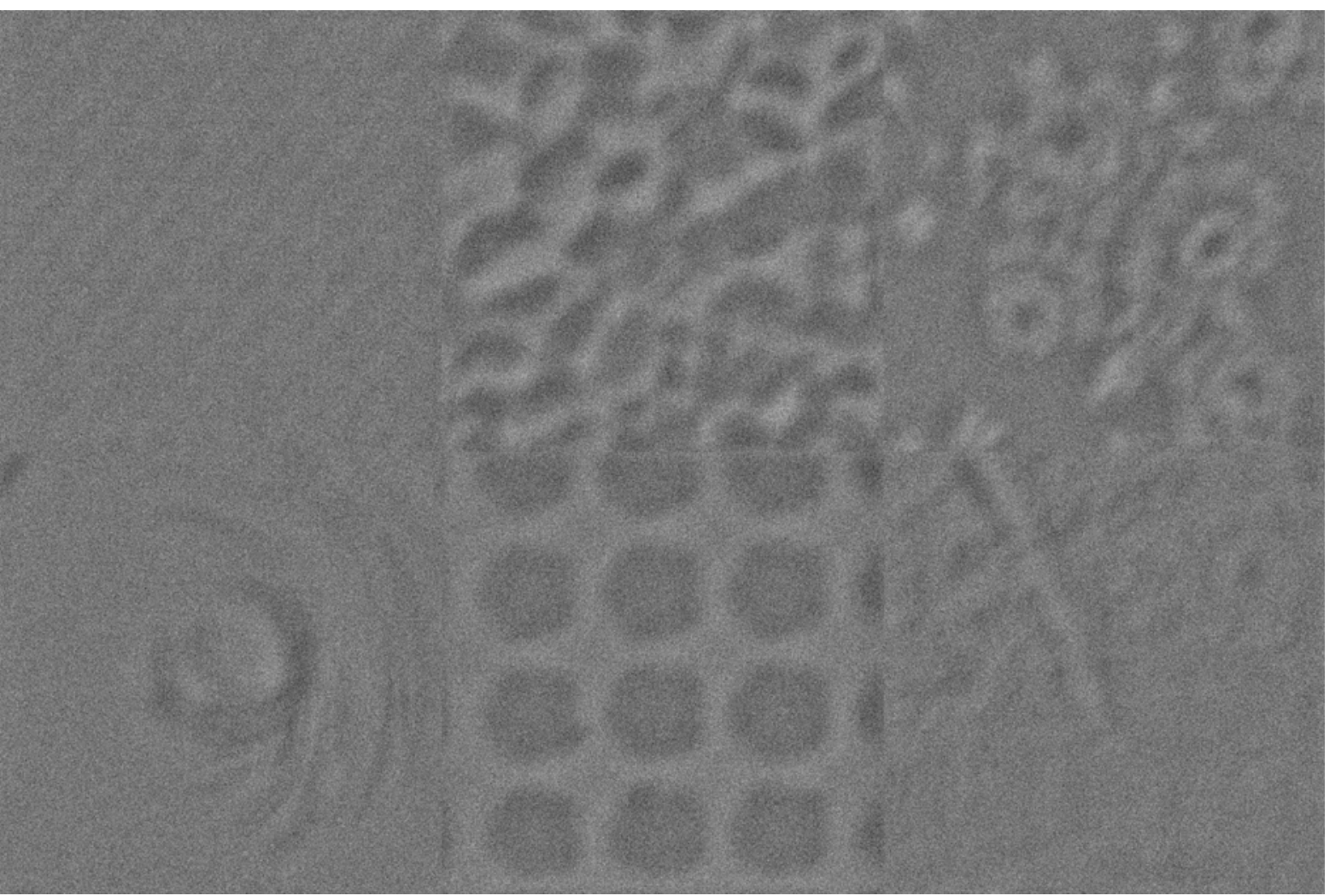}
\caption{Example of some final output images corrupted with the maximum noise level considered in this work (i.e., a SNR of -14.2 dB).}\label{fig:noise}
\end{figure}

\section{Experimental Setup and Results}\label{sect:exp}

To illustrate the potential of our scale-regularized filter learning, we set up some elementary experiments on the Brodatz image collection \cite{brodatz1966textures}.  We take the 112 images in this database as our input images (see Figure \ref{fig:in}) and corrupt them to create 112 matching $640 \times 640$ output images.  In order to do so, we first construct a kernel to convolve the original images with.
Figure \ref{fig:Forg} depicts this $640 \times 640$ filter: the gray value, the largest part of the image, takes on the value zero.  The black part has value $-1$ and the white part takes on the value $\frac{9748}{861}$, which makes sure the filter integrates to zero. Figure \ref{fig:nonoise} gives the output images after convolving the corresponding input images in Figure \ref{fig:in}.  As a final step, we add i.i.d.\ Gaussian noise to every output image.
The signal to noise ratios (SNRs) we experiment with are $65.8$ dB, $25.8$ dB, and $-14.2$ dB.  These noise levels are somewhat arbitrary, although it should be clear that if the outputs were noiseless, solving the regression without regularization would provide us with a perfect reconstruction of the original convolution kernel.   Figure \ref{fig:noise} displays the final noisy output images with the worst SNR of $-14.2$ dB, in which case we would expect the worst performance for the unregularized learning scheme.

\begin{figure}[!ht]
\centering
\includegraphics[height=4cm]{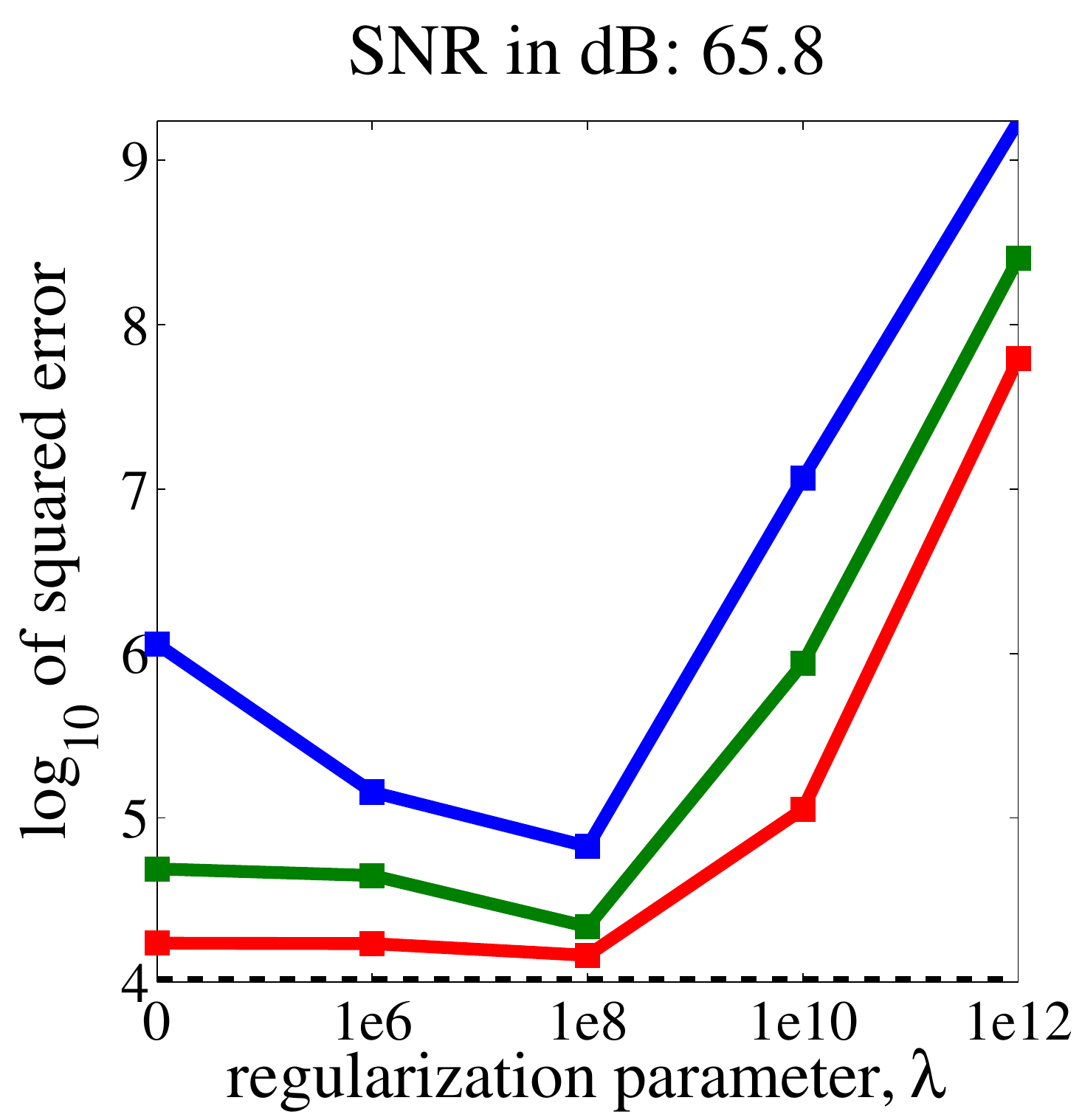}
\includegraphics[height=4cm]{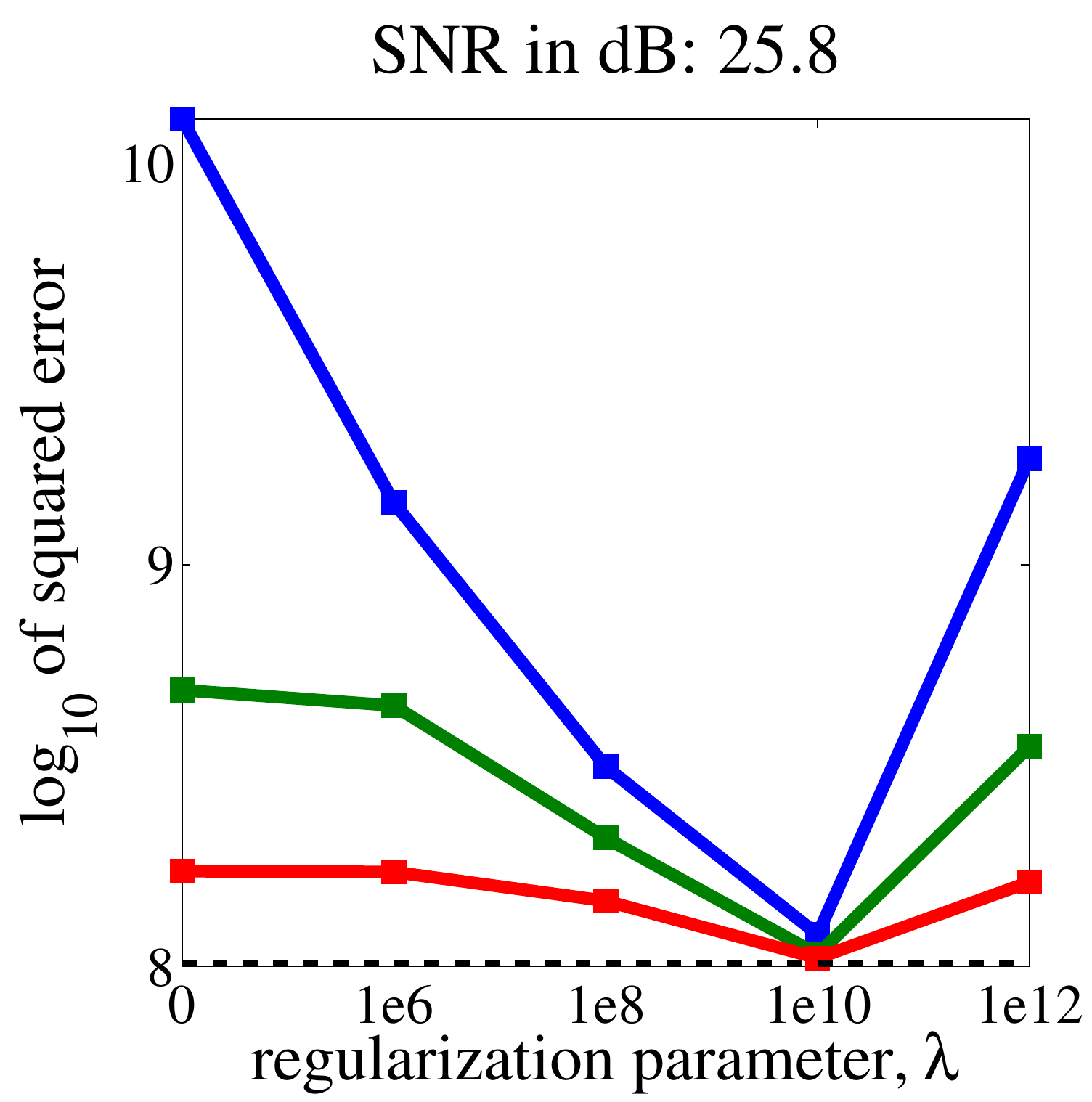}
\includegraphics[height=4cm]{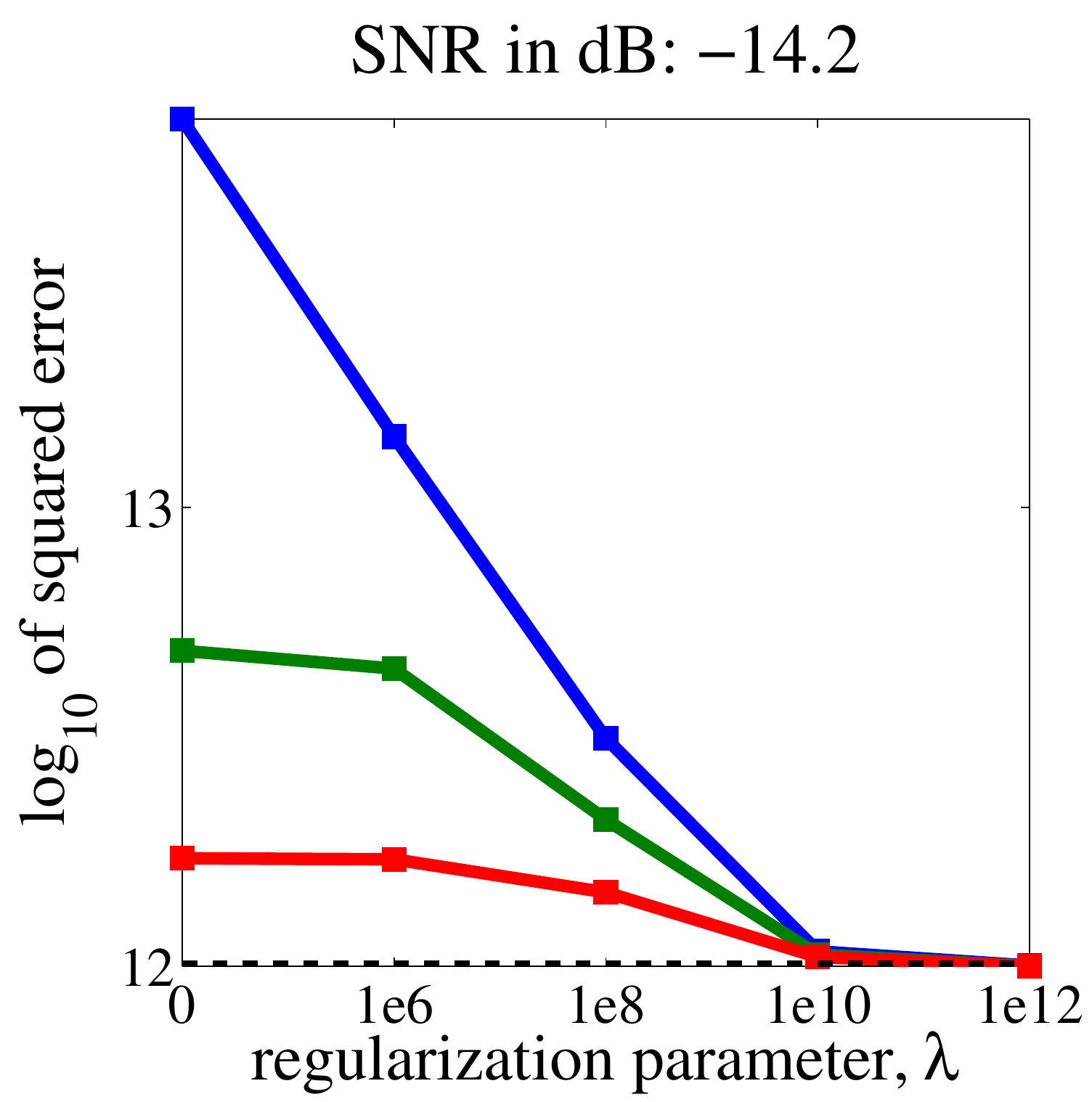}
\caption{Plots of the squared errors obtained by the nonregularized and scale-regularized regression procedures for three different noise levels (stated above every subplot). $\lambda$ is on the horizontal axis, while the vertical axis shows the 10-log of the squared error.  The blue lines are obtained when training on a single image pair; the green line on two image pairs; the red line on four.  The black, dashed line provides the error's lower bound due to noise.}\label{fig:res}
\end{figure}
In order to learn the filters, we need to decide on a test set of pairs of input and output images.  Once that has been decided, we can test the learned filters by applying them to the remaining images and measure the squared errors they achieve.  As the learning will typically improve with increasing number of images, we tested our method with learning set sizes of 1, 2, and 4 image pairs to get an impression of the behavior w.r.t.\ this aspect as well.  The values of $\lambda$ considered are $0$ (no regularization), $10^6$, $10^8$, $10^{10}$, and $10^{12}$.  Finally, as the learned filter may depend on the particular pairs of images we train on, we redid all experiments 10 times and report the averaged results.  Figure \ref{fig:res} plots the results in three subplots.  We note that all log-differences are significant according to a paired $t$-test, never giving $p$-values higher than $10^{-3}$.  The best performing regularized filter typically achieves improvements of about an order of magnitude, especially in case the learning is based on one image only (the blue lines in the plots).

\begin{figure}[!ht]
\centering
\includegraphics[height=6cm]{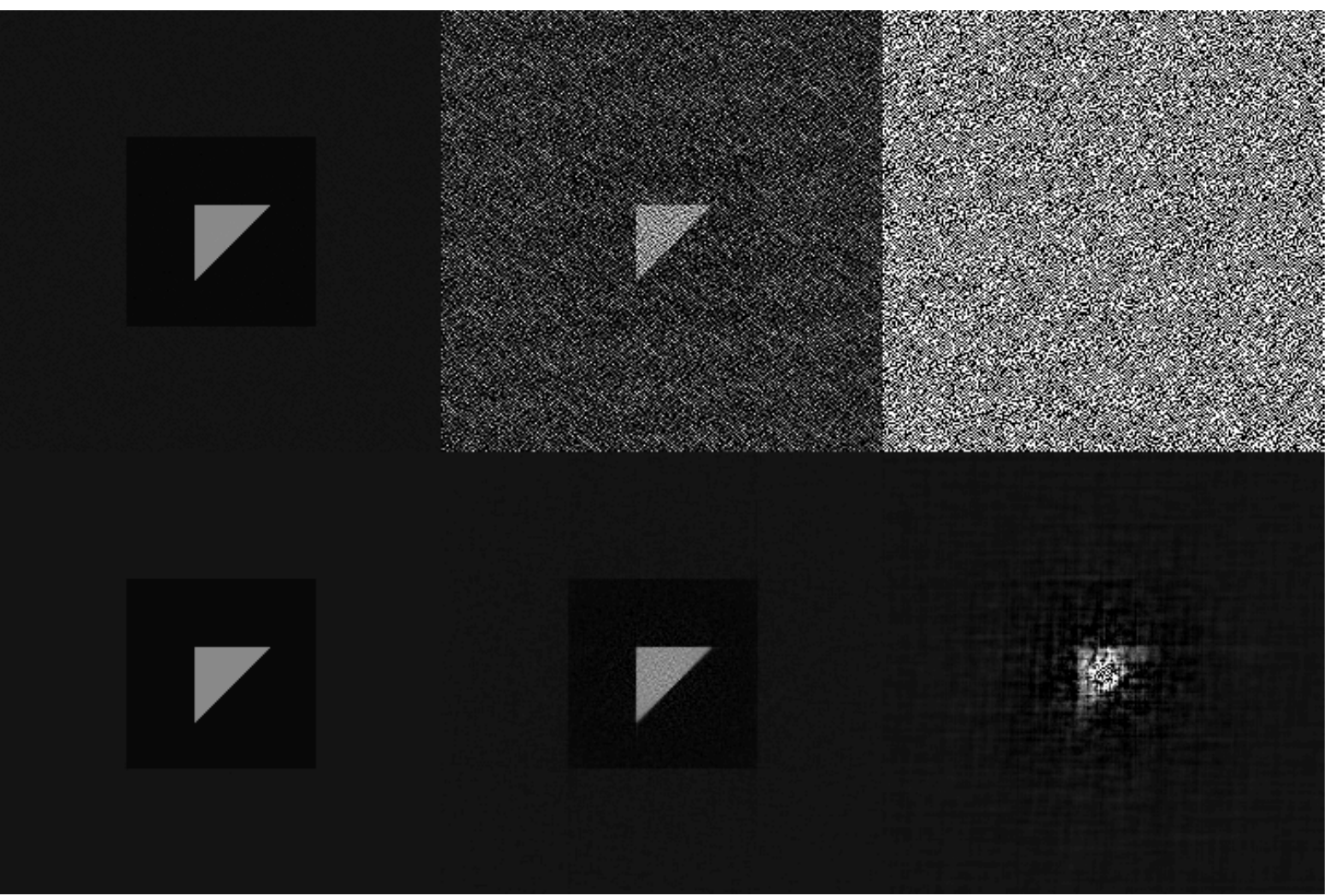}
\caption{Top row: central cropped part of the three filters estimated by means of unregularized regression.  From left to right, one filter for every one of the noise levels considered.  Bottom row: same as top row but then for the scale-regularized formulation, in which the regularization parameter is taken to be $10^8$, $10^{10}$, and $10^{12}$, respectively.}\label{fig:f}
\end{figure}
\begin{figure}[!ht]
\centering
\includegraphics[height=6cm]{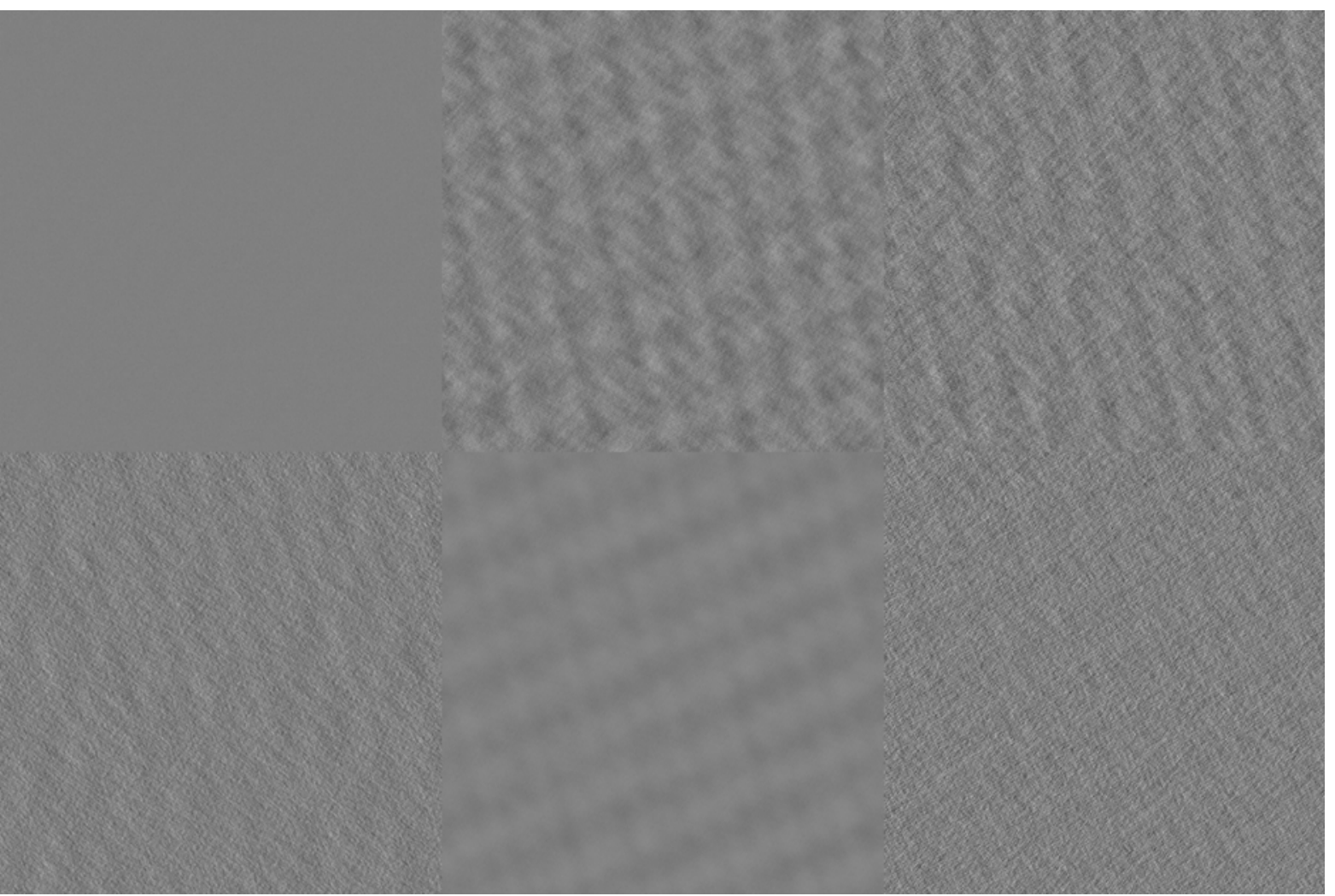}
\caption{Prediction result of unregularized regression for the images in Figure \ref{fig:in} in the setting where the maximum noise level was employed and training is based on a single image.}\label{fig:F0}
\end{figure}
\begin{figure}[!ht]
\centering
\includegraphics[height=6cm]{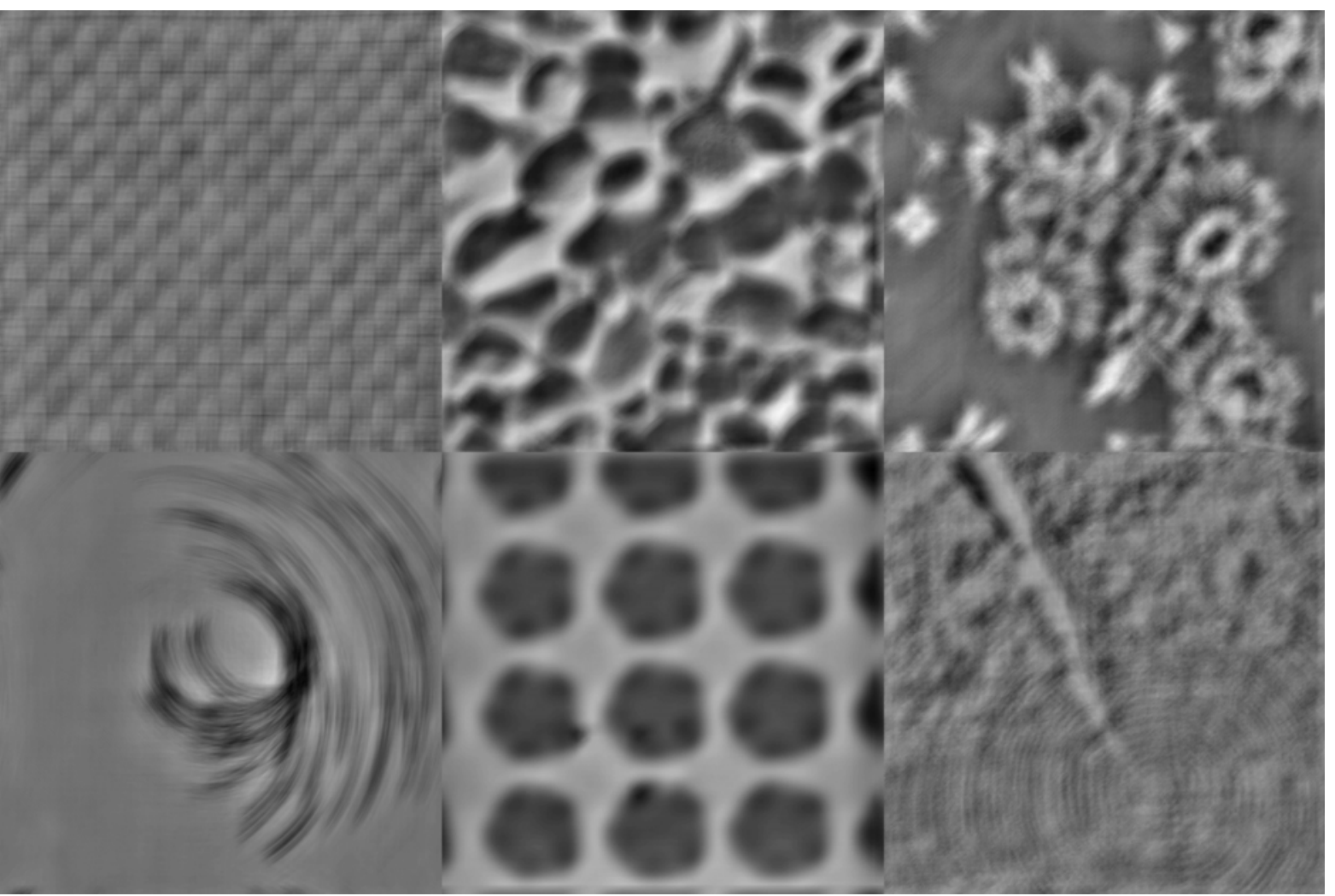}
\caption{Prediction result of scale-regularized regression for the six input images in Figure \ref{fig:in} in the setting where the maximum noise level was employed, training is based on a single image, and $\lambda$ was set to $10^{12}$.}\label{fig:Fa}
\end{figure}
For this extreme case of a single training pair, we also display some of the filters that have been inferred for the different noise levels.  Figure \ref{fig:f} shows the unregularized filters in the top row and the ``optimal'' regularized versions in the bottom row (cf. Figure \ref{fig:res}). To properly display these images, we decided to clip their gray values at twice the minimum and twice the maximum of the values attained by the original filter from Figure \ref{fig:Forg}.  For an SNR of $65.8$ dB, both procedures recover the original filter fairly well, though upon close inspection the unregularized filter clearly is more noisy in its off-center parts.  This is also reflected by the clearly inferior performance displayed in Figure \ref{fig:res}.  When the SNR reaches $-14.2$ dB, the unregularized filter values get clipped for over 98\% of the pixels, while in the case of scale-regularization this is less than 0.23\%.  This is also visually clear: while the latter, irrespective of the immense noise level, still resembles the original filter, the former has basically been reduced to noise.
Figures \ref{fig:F0} and \ref{fig:Fa} show their difference in another way and display the images obtained by convolving the input images from Figure \ref{fig:in} with the unregularized and scale-regularized kernels, respectively.  While there is little but noise visible in Figure \ref{fig:F0}, the reader hopefully appreciates the rather close resemblance---though certainly not perfect---of the reconstructed images in Figure \ref{fig:Fa} and the noiseless outputs in \ref{fig:nonoise}.


\section{Discussion and Conclusion}\label{sect:conc}

We devised a novel scheme that tackles the problem of scale in the elementary learning setting of inferring a linear filter from a set of input-output pairs.  Such a filtering is a basic building block in many a deep network, notably convolutional neural networks, that deal with signals, images, or any other input having some spatiotemporal ordering.  Our approach does not rely on any a priori restriction on the context size taken into account when performing the regression, but it incorporates a way of regulating scale by means of an added scale-regularization term that can be tuned.  Relying on variational methods from computer vision to solve the inference, it also enables us to deal with learning in spaces of very high dimensionality, which current regression method, not exploiting the underlying image structure, would be unable to solve.  In that respect, the observation that Equation \ref{eq:lin} can be optimized very efficiently is already interesting in itself.

Though some of the results are definitely quite striking already, the approach is, indeed, rather elementary.  To solve more complex, real-world filtering problems, we expect to need more complex learners.  But a basic idea of neural networks is, in fact, that one can build arbitrarily complex regression and classification schemes out of more basic building blocks.  What is essential in this, however, is that we do not have to limit ourselves to linear transformation of the data.  The next important step in this research should therefore investigate how to incorporate a so-called activation function, which transforms the filter outputs in a nonlinear way, into our setup. Introducing this nonlinearity will take us even further away from the simple-to-solve objective function in Equation \ref{eq:lin} and it is as yet unknown to what extent computational efficiency can be retained.

In the past years, contributions to top-tier venues in computer vision have been dominated by methods and solutions that reformulate the problem into pattern recognition and machine learning lingo.  We do not trivialize these achievements, but we are convinced that the proper scale space and variational methods---and the conceptual ideas underlying these methods, can further to the learning methods' impact and success.  Our contribution provides a modest step in this direction.

\bibliographystyle{unsrt}
\bibliography{supwien}

\end{document}